\documentclass[conference]{IEEEtran}
\IEEEoverridecommandlockouts
\usepackage{cite}
\usepackage{amsmath,amssymb,amsfonts}
\usepackage{algorithmic}
\usepackage{graphicx}
\usepackage{textcomp}
\usepackage{xcolor}

\usepackage{booktabs} 
\usepackage{multirow}
\usepackage{amsmath}
\usepackage{amsfonts}
\usepackage{bm}
\usepackage{dirtytalk}
\usepackage{comment}
\usepackage{url}
\usepackage{hyperref}

\definecolor{mydarkblue}{rgb}{0,0.08,0.45}
\hypersetup{colorlinks,linkcolor=red,urlcolor=mydarkblue}

\def\BibTeX{{\rm B\kern-.05em{\sc i\kern-.025em b}\kern-.08em
    T\kern-.1667em\lower.7ex\hbox{E}\kern-.125emX}}

\begin{document}

\title{Distilling Reinforcement Learning Tricks for\\Video Games}

\author{
    \IEEEauthorblockN{Anssi Kanervisto*}
    \IEEEauthorblockA{\textit{School of Computing} \\
    \textit{University of Eastern Finland}\\
    Joensuu, Finland \\
    anssk@uef.fi}
\and
    \IEEEauthorblockN{Christian Scheller*}
    \IEEEauthorblockA{\textit{Institute for Data Science} \\
    \textit{University of Applied Sciences}\\
    \textit{Northwestern Switzerland}\\
    Windisch, Switzerland \\
    christian.scheller@fhnw.ch}
\and
    \IEEEauthorblockN{Yanick Schraner*}
    \IEEEauthorblockA{\textit{Institute for Data Science} \\
    \textit{University of Applied Sciences}\\
    \textit{Northwestern Switzerland}\\
    Windisch, Switzerland \\
    yanick.schraner@fhnw.ch}
\and
    \IEEEauthorblockN{Ville Hautam\"aki}
    \IEEEauthorblockA{\textit{School of Computing} \\
    \textit{University of Eastern Finland}\\
    Joensuu, Finland \\
    villeh@uef.fi}
    
\thanks{*Equal contribution, alphabetical ordering.

\textcopyright 2021 IEEE.  Personal use of this material is permitted.  Permission from IEEE must be obtained for all other uses, in any current or future media, including reprinting/republishing this material for advertising or promotional purposes, creating new collective works, for resale or redistribution to servers or lists, or reuse of any copyrighted component of this work in other works.}
}

%\IEEEpubid{\begin{minipage}{\textwidth}\ \\[12pt]
%978-1-6654-3886-5/21/\$31.00 \copyright 2021 IEEE
%\end{minipage}}

\maketitle

\begin{abstract}
    Reinforcement learning (RL) research focuses on general solutions that can be applied across different domains. This results in methods that RL practitioners can use in almost any domain. However, recent studies often lack the engineering steps (\say{tricks}) which may be needed to effectively use RL, such as reward shaping, curriculum learning, and splitting a large task into smaller chunks. Such tricks are common, if not necessary, to achieve state-of-the-art results and win RL competitions. To ease the engineering efforts, we distill descriptions of tricks from state-of-the-art results and study how well these tricks can improve a standard deep Q-learning agent. The long-term goal of this work is to enable combining proven RL methods with domain-specific tricks by providing a unified software framework and accompanying insights in multiple domains.
\end{abstract}


\section{Introduction}
    
    Reinforcement learning (RL) is a form of machine learning that builds on the idea of learning by interaction and learns solely from a numerical reward signal. Due to its generality, reinforcement learning has found applications in many disciplines and achieved breakthroughs on many complex tasks, including modern video games \cite{vinyals2019grandmaster, openai2019dota}.
    
    Alas, such feats often require a large amount of training data to learn. For example, in the case of the \say{OpenAI Five} agent, it took roughly $40,000$ years of in-game time to train the agent~\cite{openai2019dota}. Instead of more training, one can incorporate human domain-knowledge of the task and the environment to aid the training of the agent (\say{tricks}), which are especially prominent in the RL competitions~\cite{vizdoom_competitions, unity2019obstacle, milani2020minerl}. As these tricks are domain-specific, they are rarely covered by academic publications. Nevertheless, their prevalence across domains hints that they are a necessary step in practical applications. If a trick works, it also tells something about the structure of the task and how it could be solved in a more general fashion.
    
    To broaden the knowledge on these tricks, we summarize the approaches used by participants in RL competitions and state-of-the-art results. We categorize these tricks and then apply them on three complex RL environments to conduct preliminary ablation experiments to study how they affect performance.

    \begin{figure}[t]
        \centering
        \includegraphics[width=0.98\columnwidth]{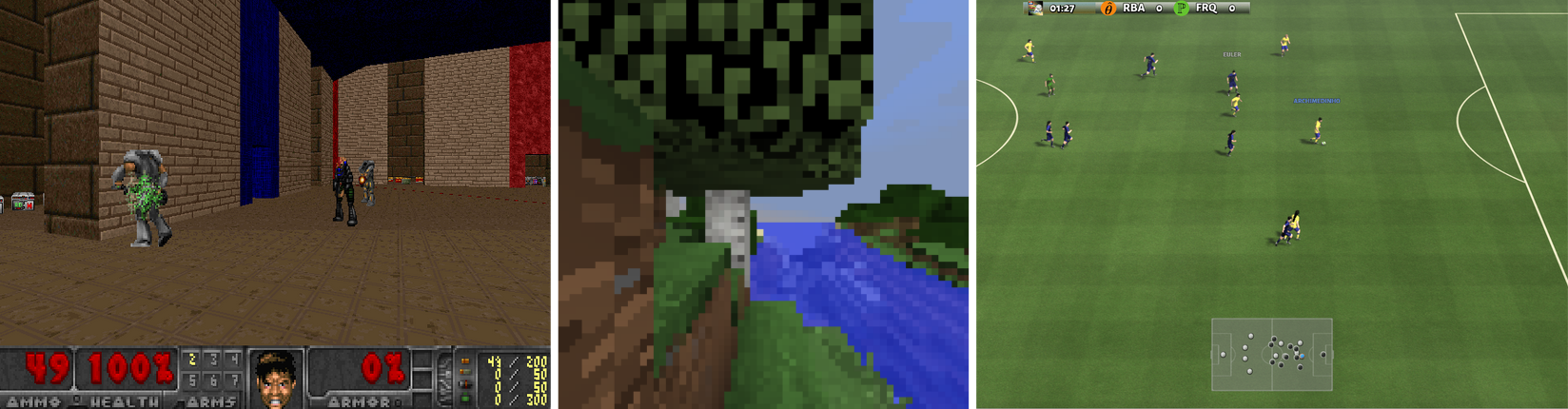}
        \caption{Environments used in the experiments: ViZDoom Deathmatch, MineRL ObtainDiamond, and Google Research Football environment, in this order. Pictures of ViZDoom and MineRL depict image inputs the agent receives, while in Football agent receives direct information, such as the location of the players.}
        \label{fig:envs}
    \end{figure}
    
    % This is costly and prohibits its application in domains where large amount of samples are not easily obtained.
    % Introducing modifications guided by human heuristics (tricks) can significantly speed up the learning [examples, cite]
    % Since these modifications (tricks) are usually domain specific, they do not receive a lot of attention from research, where the focuses mostly lies on general solutions.
    % However, they find wide-spread use in RL competitions [cite them all], super-human agents [cite dota, sc2] and applications [cite game-testing], where they often play a crucial part.
    % Since the tricks are commonly domain specific, they have to be adjusted or re-implemented for a new problem at hand. 
    % This calls for a framework that supports this process by abstracting away common stuff and guiding the developer.
    
    % In this work we first summarize such tricks used in ....
    % We then come up with a unified framework, that unites these tricks and that allows fast adaptation to new problems.
    % We then apply this framework to three video-game domains: Doom, MineCraft and Google football [cite them].
    % For each of these domains, we implement and compare different domain-specific tricks.
    
    %\todo{Remember to keep focus on video games}
    
    %\todo{Cite \cite{bergdahl2020augmenting}, they do "scripting with RL", we do "RL with scripting"}

\section{Reinforcement learning tricks in the wild}
    \label{sec:tricks}
    
    \begin{table*}[t]
\caption{Summary of domain-specific methods used to support the training of RL agents.}
\centering
\begin{tabular}{lll}
\toprule
\textbf{Trick name and abbrevation} & \textbf{Description} & \textbf{References} \\ \midrule
Reward sharping (RS)                                    & Modify reward function for a denser reward & \cite{lample2017playing, vizdoom_competitions, jaderberg2019human, vinyals2019grandmaster, Wu2017Doom, alex2019competing, unity2019obstacle, openai2019dota}                                        \\
Manual curriculum learning (CL)                                & Increase difficulty gradually             &   \cite{dosovitskiy2016learning, Wu2017Doom, lample2017playing, vizdoom_competitions, clemens2021codecraft, openai2019dota, Rychlicki2020ZPPGFootball} \\
Manual hierarchy (MH)                                   & Split task into subtasks by manual rules & \cite{lample2017playing, vizdoom_competitions, unity2019obstacle, song2019playing, patil2020align, skrynnik2020forgetful, bergdahl2020augmenting}                                         \\
Modified actions (MA)                                   & Modify agent's actions while learning & \cite{lample2017playing, vizdoom_competitions}                                        \\
Scripted actions (SA)                      & Augment the agent with hardcoded actions & \cite{Wu2017Doom, vizdoom_competitions, openai2019dota}                                        \\
%Imitation Learning (IL)                                 & Pretrain agent on expert demonstrations & \cite{vinyals2019grandmaster, vizdoom_competitions, alex2019competing, milani2020minerl} \\
%Omniscient value function                               & Value function is given more information than for the policy        & \cite{vinyals2019grandmaster, clemens2021codecraft}                                         \\
%Exploring starts                                           & Randomly start agent mid-way into the episode                  &  \cite{patil2020align}                                      \\
\bottomrule
\end{tabular}
\label{table:summary}
\end{table*}
    
    In this work, a \say{trick} refers to a technique outside the RL algorithm that improves the training results. We treat the RL algorithm as a black-box function that takes in observations and outputs actions, and given the reward signal learns to pick actions with higher rewards. Tricks in this work take this black-box function and build up around it to create an agent. This means we do not include the choice of RL algorithm, code-level improvements (e.g. normalizing values), or network architecture choices, which can lead to improvements~\cite{vinyals2019grandmaster, clemens2021codecraft}. We also do not include observation or action space shaping, as these have been explored in previous work~\cite{kanervisto2020action}. We emphasize that we are interested in any reported tricks which improve the results, regardless of the lack of generalizability.
    
    Table~\ref{table:summary} categorizes the tricks described in the reviewed works. Some of these are common knowledge, like reward shaping (RS)~\cite{ng1999policy}, but others are less explored. For example, Lample et al. (2017) \cite{lample2017playing} assist the learning agent by modifying the actions (MA) to move the crosshair on top of the enemies before shooting, but not during evaluation. This action is added to the agent's learning buffers, from which it learns to shoot by itself. Another common trick is to hardcode the agent's behavior with scripted actions (SA), which can support the agent with easy-to-script solutions~\cite{Wu2017Doom}. Augmenting scripted behavior with RL has also been shown to be an effective strategy~\cite{bergdahl2020augmenting}. Another common solution to sparse reward environments is to use curriculum learning to initially start with an easy problem and gradually increase the difficulty~\cite{bengio2009curriculum}. This can be done by increasing the health and difficulty of the opponents~\cite{Wu2017Doom} or by gradually adding more tools to the agent's disposal~\cite{clemens2021codecraft}. 
    
    %\todo{why would you do SA than MA? MA is useful if you want to train the agent to do stuff like learning to detect something from an RGB image without a help of a labels buffer. SA is good if you can do the scripting based on the information you already have (e.g. in minerl crafting you do not need additional information)}
    
    Sometimes the task can be split into multiple subtasks, and assigning one learning agent to each has yielded significant performance gains~\cite{vizdoom_competitions, song2019playing, bergdahl2020augmenting}. This is credited to the simplicity of the subtasks (easier for agents to learn) and to catastrophic forgetting, where learning skills for new tasks might interfere with skills learned for a previously learned task. In most cases, one or more of these tricks were combined into one. For example, a scripted logic decides which of the subtask agents should take control next~\cite{vizdoom_competitions}.
    
    %\todo{Mention IL, omniscient value function and exploring starts as interesting future things}

\section{Experimental setup}
    With the above categorization of the tricks, we pick the most prominent ones, implement them along with a deep Q-learning (DQN)~\cite{dqn} RL agent and study their performance in three different domains: Doom, Minecraft, and a Football environment. We chose these environments as we, the authors, are experienced in using these environments, and are familiar with their challenges. We spread experiments over multiple environments to gain more general insights rather than domain-specific knowledge. 
    
    We choose DQN as it is an off-policy RL algorithm, meaning we can modify the actions for learning. Also, DQN's performance in discrete action spaces is well documented and reputable implementations of it exist. We use the implementation from the stable-baselines3 repository~\cite{stable-baselines3}. The source code used for the experiments is available at \url{https://github.com/Miffyli/rl-human-prior-tricks}.
    
    \subsection{Reinforcement learning agent}   
        By default, the DQN agents have a replay buffer of one million transitions, start learning after 10K steps and update the target network after 10K training steps. If the agent input is an RGB image, it is processed with a convolutional layer network of the original DQN paper~\cite{dqn}. The agent is updated at every step to emphasize sample-efficient learning. For exploration, we use an $\epsilon$-greedy strategy, where agents pick a random action with a small probability. In ViZDoom and MineRL the chance of a random action is linearly decayed from 100\% to 5\%/10\% in the first 10\%/1\% training steps, respectively. In GFootball the probability is fixed at 1\%. Other settings are taken from default settings of stable-baselines3~\cite{stable-baselines3}.
    
    \subsection{ViZDoom Deathmatch}
        The ViZDoom learning environment~\cite{vizdoom} provides an interface to train RL agents in the Doom environment, based on visual input (image pixels) similar to what a human player would see (see Fig.~\ref{fig:envs} for an example). We use the \say{Deathmatch} scenario where the agent is rewarded for killing continuously spawning enemies. The scenario also includes weapons and pickups for health and armor. Since killing an enemy takes multiple shots, especially with the starting weapon (a pistol), the reward is sparse, and with a crude exploration strategy the RL agent has a hard time learning the task (see results in Section~\ref{sec:results}). 
        
        The same challenges arose in the ViZDoom competitions~\cite{vizdoom_competitions}, where instead of game enemies, the agents had to play against each other. Competition participants employed various tactics, most notably MH of agents~\cite{lample2017playing, song2019playing}, CL by starting with weaker enemies~\cite{Wu2017Doom} and modifying agent's actions (MA) to optimal ones during training~\cite{lample2017playing}. When combined and tuned correctly, this turns basic RL algorithms into capable agents in these competitions, approaching human-level performance but not quite surpassing it~\cite{vizdoom_competitions}. 
        
        We implement three of these tricks and try them in the deathmatch scenario: RS, MA, and MH. With RS, the agent is given a small reward (0.1) for hitting an enemy or when picking up pick-ups. With MA, we force the agent to shoot if an enemy is under the crosshair (used only during training). In MH we create two learning agents, one for navigation and one for combat, where the combat agent is activated when an enemy is visible on the screen, otherwise, the navigation agent plays. These two agents are separated with their own replay buffers for training. When combined with RS and MA, both agents use the same RS and the shooting agent also uses MA.
        
        Agents are provided with an RGB image of size 80x60, and on each step can choose one of 66 actions, which consist of different combinations of buttons allowed (turning left/right, moving forward/backward/left/right, shooting, selecting next/previous weapon, turning 180 degrees, and enabling sprinting). The agent chooses an action every four environment steps. During the evaluation, we always pick the action with the highest Q-value.
    
    \subsection{MineRL ObtainDiamond}
    
        \begin{figure}[t]
            \centering
            \includegraphics[width=0.98\columnwidth]{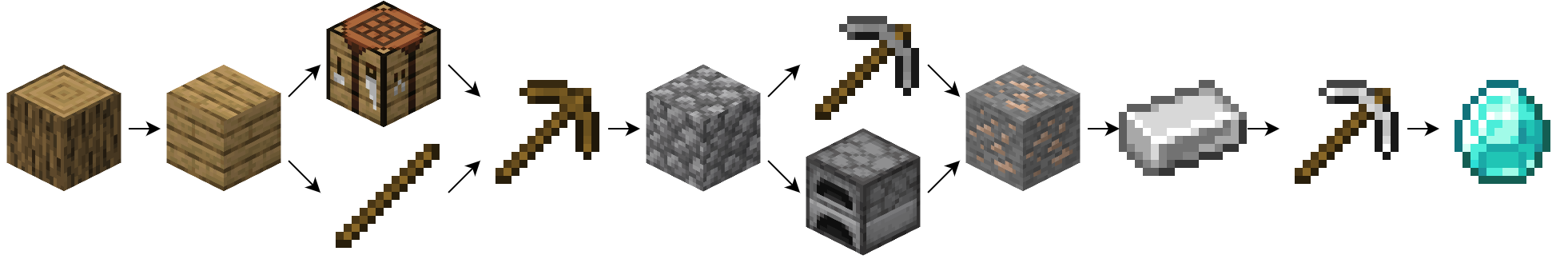}
            \caption{MineRL ObtainDiamond item hierarchy. Subsequent items yield exponentially higher rewards.}
            \label{fig:obtain_diamond_items}
        \end{figure}
    
        MineRL ObtainDiamond \cite{mineRLcompetition} is an environment built in Minecraft, where the goal is to obtain a diamond.
        The player starts with no item in a randomly generated level.
        To mine a diamond, the player must first collect and craft a list of prerequisite items that hierarchically depend on each other (see Fig.~\ref{fig:obtain_diamond_items}).
        The agent is rewarded the first time they obtain an item in this hierarchy tree.
        The game ends when the player obtains a diamond, dies, or after a maximum of 15 minutes in-game time.
        
        The MineRL ObtainDiamond task poses multiple challenges for reinforcement learning agents.
        The sparsity of the reward signal and the complexity of the state- and actions-space make it a particularly hard exploration problem.
        Furthermore, random level generation means that agents must generalize across a virtually infinite number of levels.
        %Additionally, the simulator is slow compared to other RL video game benchmarks, which makes scaling to large numbers of training steps costly.
        
        To address these challenges, we use three tricks: RS, MH, and SA.
        With RS, the agent receives a reward each time a needed item is obtained, until the required amount is reached. 
        This is opposed to the default reward function that only awards an item once when it is collected for the first time. 
        %It has been shown, that hierarchical reinforcement learning is a promising step towards strong performing agents on MineRL ObtainDiamond~\cite{milani2020minerl}.
        As for MH, we train one agent per intermediate item. This type of hierarchical solution has had high performance in the MineRL competitions~\cite{milani2020minerl}.
        Finally, some of the items require a specific sequence of crafting actions, which can be easily scripted.
        With SA, we replace sub-policies for crafting items with fixed policies that follow a pre-defined action sequence. 
        
        Agents perceive the environment through an RGB render image of size 64x64 and a list of the items currently equipped and in the inventory.
        We simplify the action space of all agents following~\cite{kanervisto2020action}, by discretizing camera actions, dropping unnecessary actions (these vary between agents) and flattening multi-discrete actions to discrete actions. Agent picks an action once every eight frames.
        %We trained each agent for a total of 2.5 million environment frames (roughly 3.5 hours in-game time), whereas a frame-skip of 8 was used.
        % Final agents were evaluated on 200 episodes to obtain an estimate of the expected episodic reward.
        
    \subsection{The Football Project (GFootball)}
        Google Research Football~\cite{kurach2020google} is a 3D RL environment that provides a stochastic and open-source simulation.
        % As a multiplayer game, the environment facilitates research of the effects of self-play, where the agent plays against different versions of itself.
        The engine implements an 11 versus 11 football game with the standard rules, including goal kicks, free kicks, corner kicks, yellow and red cards, offsides, handballs, and penalty kicks. Each game lasts 3000 frames.
        In the 11 versus 11 football game, the RL agent faces a built-in scripted agent whose skill level can be adjusted with a difficulty parameter between 0 and 1.
        The agent controls one player at a time, the one in possession of the ball if attacking or the one closest to the ball if defending.
        % Players have different characteristics like speed or accuracy, but both teams have the same set of players.
        % Furthermore, players are getting tired over time, which influences their behavior and skills.
        The agent is rewarded with +1 when scoring and -1 when conceding a goal.
        
        Playing football is challenging as it requires a balance of short-term control, learned concepts such as tackling, and high-level strategy.
        The agents face a sparse reward signal, complex action- and observation-spaces, and a stochastic environment.
        
        We approach these challenges in the 11 versus 11 easy stochastic environment with two tricks: MH and CL.
        With MH we evaluate a simple manual hierarchy with separate sub-policies for situations with and without ball possession. This allows each sub-policy to specialize in attacking or defending independently.
        In CL we gradually increase the game difficulty parameter as described in~\cite{Rychlicki2020ZPPGFootball}. The agents receive a 115-dimensional vector summarizing the game state. The agent can choose one of 19 actions consisting of different football-specific movements like passing, tackling, shooting, and dribbling at every step. 
        % We evaluated the impact of those two tricks on the 11 versus 11 easy stochastic environment isolated and combined.
        %The training ends after 10 million frames.

\section{Results and discussion}
    \label{sec:results}
    \begin{figure*}[t]
        \centering
        \includegraphics[width=1.98\columnwidth]{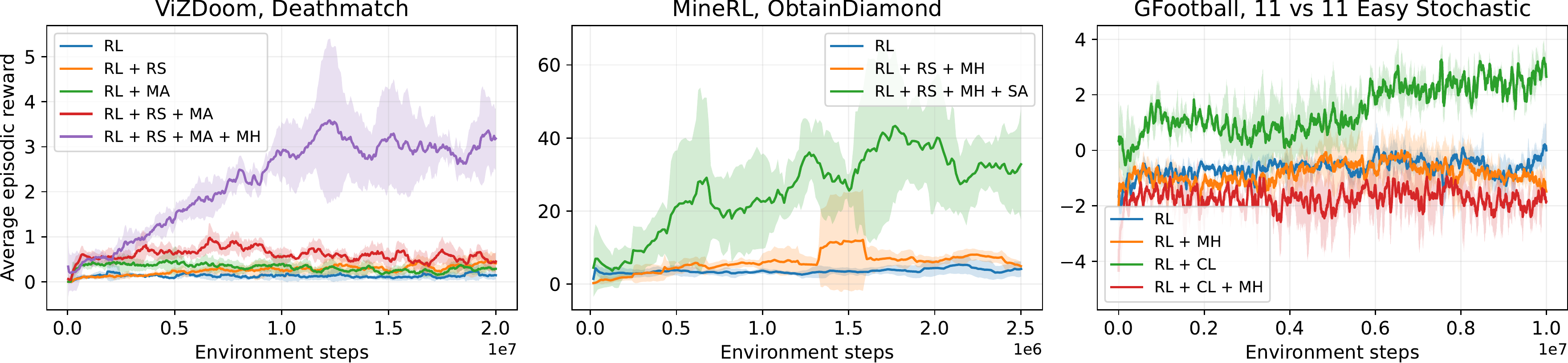}
        \caption{Training results with different tricks on top of a normal DQN agent (RL). Abbreviations are as in Table~\ref{table:summary}. RL is a vanilla DQN learning agent without tricks. We average results over five training runs for ViZDoom and MineRL, and three for GFootball.}
        \label{fig:main-results}
    \end{figure*}
        
    \begin{table}[t]
      \caption{Final performances of the MineRL agents, evaluated on 200 episodes, and averaged over five training runs. ``Best" is the average performance of the best agent across the five runs. ``Max" is the highest per-episode score reached while evaluating the agents from the five runs.}
      \small
      \centering
        \begin{tabular}{lccc}
        \toprule
        \textbf{Experiment} & \textbf{Mean} & \textbf{Best} & \textbf{Max} \\
        \midrule
        RL & $3.8 \pm 1.0$ & $5.4$ & $35$ \\
        RL+RS+MH & $4.5 \pm 0.7$ & $5.4$ & $99$ \\
        RL+RS+MH+SA & $33.5 \pm 5.6$ & $41.4$ & $547$ \\
        \bottomrule
      \end{tabular}
      \label{resultTable}
    \end{table}
    
    The learning curves are reported in Fig. \ref{fig:main-results}. In ViZDoom, all tricks improve upon the DQN, but only RS, MA and MH combined provide significant improvements. Most notably, MH from an average score of less than one to an average score of three, despite only splitting the task into two separate learning agents. 
    
    In the MineRL experiments, MH and RS only slightly improved the expected mean reward compared to DQN.
    However, the maximum achieved reward increased from 35 to 99 (Table \ref{resultTable}), which corresponds to two steps in the item hierarchy.
    SA in combination with RS and MH resulted in the significantly fastest training and best final performance.
    
    In the GFootball environment, Only CL improved the mean reward over the baseline.
    The use of MH isolated and in combination with CL did not improve the average score, in the latter case, it even led to decreased performance.
    When using MH the attacking and defending agents have their separate neural networks with no weight sharing.
    Therefore, each network has fewer updates than the DQN or CL agents, which could cause the performance drop.
    One could mitigate this problem by implementing the agents with weight sharing and their independent value heads.
    
    %In   summary,   most   of   the   evaluated   tricks   did   lead   toimprovements compared to DQN, albeit their impact was dif-ferent depending on the environment
    In summary, most of the evaluated tricks did lead to improvements over vanilla DQN, with a right combination yielding significantly better results.
    Out of the individual tricks, manually separating the task into subtasks (MH) yielded the most promising results so far, but with a negative impact on the GFootball environment.
    This concurs with the original observation that, while these tricks are likely beneficial, the exact effect is highly dependent on the environment.
    %An exception was MH, which improved the performance on VizDoom and MineRL ObtainDiamond, but truned out to be harmfull on 11 versus 11 Google football.
    %These results highlight the potential but also the domain-specificity of these tricks.

    % The benchmark \cite{kurach2020google} Ape-X DQN reached an average reward of $-1.15 \pm 0.37$ after 20 million frames on the 11 vs 11 easy stochastic environment.
    % The effects of the hacks applied on this environment are difficult to judge. 
    % After only 10 million frames, we can surpass the benchmark results.
    % Curriculum learning by gradually increasing the difficulty stabilizes training.
    % To gain more insights into our hacks' effects, we need to run the experiments with five different random seeds.
        
\section{Conclusions and future work}
    In this work, we summarized the different domain-knowledge techniques (\say{tricks}) used by researchers in state-of-the-art results, both in academic publications and competitions. While individual tricks rarely generalize outside the specific environment and task, we find that tricks of the same category (e.g. reward shaping, hierarchy) span various environments and, overall, are reported as significant steps to obtain the reward. We then took the most prominent tricks, implemented them in three different environments along with a DQN agent, and ran preliminary experiments to study their impact on the learning. The results so far indicate that manually splitting the task into subtasks and hardcoding the agent's actions are very effective tricks to improve the agent's performance, especially when these are easy to implement.
    
    The future of this work consists of more systematic study and categorization of the tricks, providing more formal definitions for the tricks to understand them better, and finally conducting large-scale experiments over various environments for general information. Along with this, a unified software framework for combining scripted and learned behavior would be a useful tool for easing the use of RL in practical applications. Finally, even though the tricks assume human domain knowledge is available, we believe these large-scale results on them would serve as a useful tool for RL applications and a source of directions for RL researchers.

\bibliographystyle{ieeetr}
\bibliography{main}

\vspace{12pt}

\end{document}